\documentclass[runningheads]{llncs}
\usepackage{graphicx}
\begin{document}
\title{Predicting Participants' Performance in Programming Contests using Deep Learning Techniques\vspace{-3ex}}
\titlerunning{Predicting Participants' Performance in Programming Contests}
%
\author{Md. Mahbubur Rahman\inst{1}\and
Badhan Chandra Das\inst{2} \and
Al Amin Biswas\inst{3}
\and 
Md. Musfique Anwar\inst{2}\vspace{-2ex}}

\authorrunning{Rahman et al.}
%
\vspace{-5ex}
\institute{Iowa State University, Ames, Iowa, USA, \and
Jahangirnagar University, Dhaka, Bangladesh, \and
Daffodil International University, Dhaka, Bangladesh.
\email{}\\
\url{} 
\email{mdrahman@iastate.edu, badhan0951@gmail.com, \\alaminbiswas.cse@gmail.com, manwar@juniv.edu }}
\maketitle              
\vspace{-5ex}
\begin{abstract}
In recent days, the number of technology enthusiasts is increasing day by day with the prevalence of technological products and easy access to the internet. Similarly, the amount of people working behind this rapid development is rising tremendously. Computer programmers consist of a large portion of those tech-savvy people. Codeforces, an online programming and contest hosting platform used by many competitive programmers worldwide. It is regarded as one of the most standardized platforms for practicing programming problems and participate in programming contests. In this research, we propose a framework that predicts the performance of any particular contestant in the upcoming competitions as well as predicts the rating after that contest based on their practice and the performance of their previous contests.
\vspace{-2ex}
\vspace{-2ex}
\end{abstract}
\vspace{-5ex}
\section{Introduction}
\vspace{-2ex}
Codeforces is an online programming practice and contest hosting platform maintained by a group of competitive programmers from ITMO University, led by Mikhail Mirzayanov. According to Wikipedia, there were more than 600,000 registered users on this site. There are several certain features of Codeforces as follows. 
This site has been developed specially for competitive programmers while preparing for the programming contests. A registered user of this platform can use it in terms of practicing anytime and participating in the contests running at that time with the facility of the internet. There is a rating system commonly known as divisions of each contestant taking part in the contests based on their performance, i.e. capability to solve the problems according to their difficulty level of that contest as well as the previous ones. The rating system, divisions and titles are shown in Table \ref{tab:rating_table}. The contestants can try to solve the unsolved problems of any contests, even after the contest, also known as upsolve. There are several types of contests that can be hosted in Codeforces. Among them, the most popular one is short contests held for two hours, which is also known as Codeforces Round. It can be conducted once a week. Another one is a team contest, where any registered user can invite any other registered users (at most two) for a contest. The users can also get connected (follow- following) with each other in order to watch updates of them. The trainers or institutions who organize the contests usually do this to track the progress of the trainees and students.
One of the important and effective features of this widely used platform is, there is a community platform like Stack overflow, to get the solutions of the problems faced during the contest and in practice. However, this difference between this community platform and others is, it is dedicated for the competitive programmers trying to solve any programming problems while practicing independently or the problems after the contests. The users can also get a list of tagged problems, e.g. dynamic programming problems, greedy problems, etc. to practice and get experts or work on the weak parts of him or her on specific types of problems.
\vspace{-6ex}
\begin{table}[!ht]
\centering
\label{tab:rating_table}
\caption{Codeforces User Rating and Divisions}
\vspace{-1ex}
\begin{tabular}{|c||c||c||c|}

\hline
\textbf{Rating Bounds} & \textbf{Color} & \textbf{Division} & \textbf{Title}                                                       \\ \hline \hline
\textgreater{}= 3000   & Black \& Red   & 1                 & \begin{tabular}[c]{@{}c@{}}Legendary\\  Grandmaster\end{tabular}     \\ \hline
2600 -- 2999           & Red            & 1                 & \begin{tabular}[c]{@{}c@{}}International \\ Grandmaster\end{tabular} \\ \hline
2400 -- 2599           & Red            & 1                 & Grandmaster                                                          \\ \hline
2300 -- 2399           & Orange         & 1                 & \begin{tabular}[c]{@{}c@{}}International \\ Master\end{tabular}      \\ \hline
2100 -- 2299           & Orange         & 1                 & Master                                                               \\ \hline
1900 -- 2099           & Violet         & 1/2               & \begin{tabular}[c]{@{}c@{}}Candidate \\ Master\end{tabular}          \\ \hline
1600 -- 1899           & Blue           & 2                 & Expert                                                               \\ \hline
1400 -- 1399           & Cyan           & 2/3               & Specialist                                                           \\ \hline
1200-1399              & Green          & 2/3               & Pupil                                                                \\ \hline
\textless{}= 1199      & Gray           & 2/3               & Newbie                                                               \\ \hline
\end{tabular}
\end{table}
\vspace{-5ex}

In this research, we propose a framework which predicts the performance of each individual programmer in upcoming contests based on his or her previous contests. The performance of the contestants is performed in two perspectives. First, we predict whether the rating of the contestant will increase or decrease, second, the rating itself of that corresponding contestant. The performance tracking of the contestants in order to recommend him to improve his performance in the impending ones. The main contributions of this paper is as follows.

\begin{enumerate}

    \item We predict the performance of each contestant by analyzing his or her performances in the previous contests and practice problems.
    \item The ratings of the programmers will also be predicted along with the percentage of increase or decrease of their ratings.
    
    \item This experimental research is conducted on a real-world dataset obtained from Codeforces.

\end{enumerate}

\vspace{-1ex}
The remaining sections are organized as follows. The section 2 covers relevant works in this topic. In sections 3 and section 4, we explain the problem definition and proposed methodology respectively.The experimental outcomes are presented in section 5. In section 6, we conclude the paper.

\vspace{-2ex}
\section{Literature Review}
\vspace{-2ex}
To identify the gap in the available research, we have conducted extensive searches and investigations of numerous related studies. However, a very little amount of research work has been accomplished on this topic. 
Using the students’ data of secondary school, Amra et. al \cite{Amra} applied KNN and Naive Bayes classifiers to predict the students' performance. The obtained result showed that Naive Bayes outperformed KNN by attaining the accuracy of 93.6\%. Babić et. al. \cite{Ivana} tried to imbed the links between student academic motivation and their behaviour in the learning management system (LMS) course. Three different machine learning (ML) classifiers namely neural networks, support vector machines, and decision trees were applied to classify the students. Though the performance of all the classifiers were significant but the neural network was more promising than others applied models in detecting the student academic motivation based on the behaviour.
\vspace{-3ex}
\subsection{Academic Performance Prediction}
Waheed et al. attempted to develop a system that can predict students' academic success based on clickstream data and assessment results in a virtual learning environment. They used the artificial neural network (ANN) to classify the student performance into different classes and compared the obtained result of ANN with two baseline methods namely support vector machines and logistic regression \cite{Waheed}. It is observed that ANN outperformed the baseline methods.
Several works related to student performance prediction have also been accomplished \cite{Xu}, \cite{Al-Shabandar},\cite{Zulfiker},\cite{Ofori}. 
\vspace{-2ex}
\subsection{Contest Performance Prediction}
Sudha et al. \cite{Sudha} worked on the classification and recommendation of competitive programming problems using Convolution Neural Network (CNN). The goal of their proposed system is to determine the required approach for solving the problem.
W. Looi  analyzed single C++ source code submission on Codeforces and tried to predict a user’s rank and country \cite{Looi}. Among all the applied models, the neural network attained the highest accuracies of 77.2\% in rank prediction (within one rank) and 72.5\% in the country's prediction. A. Alnahhas et al. investigated ML techniques to develop a system that can predict the contestant's future performance by dissecting their past rating record \cite{ALNAHHAS}. Here, they applied five different baseline machine learning approaches. Besides this, they proposed a new deep learning model for result comparison with baseline. To conduct this research, they collected public data from the Codeforces website. They found that most of the applied techniques attain an acceptable result but the deep learning model performed better than the baseline.
Chowdhury et al. \cite{Intisar} trained a Kohonen Self organizing feature map (KSOFM) neural network on log data regarding programmers’ performance. Here, programmers are grouped into three distinct clusters ie. ‘at risk’, ‘intermediate’, and ‘expert’. The proportional rules made classification with an accuracy of 94.00\%. Besides this, three more models namely multilayer neural networks, decision tree, and support vector machine were trained using the same dataset.  Among them, feedforward multi-layer neural networks and decision trees have achieved an accuracy of 97.00\% and 96.00\% respectively. The precision of the support vector machine was about 88.00\%, but it attained the highest recall of 99.00\% in terms of distinguishing ‘at risk’ students.
By investigating ten years of TopCoder algorithm competitions, J. R. Garciaa et al. reported on the learning curves \cite{Garcia}. They also discussed how these learning curves are employed in university courses. Later, it can aid them to explain the impact of competitive programming in a class.\\
Ishizue et al. \cite{Ishizue} employed machine learning models to try to simplify the process of predicting placement outcomes outside of the conventional, time-consuming placement examination and the level of programming competence outside of a programming contest. The explanatory variables consist of psychological assessments, programming tasks, and student-completed surveys.

Ohashi et al. proposed a unique feature extraction technique and convolutional neural networks to classify the source code. To  demonstrate the proposed algorithm, they have used data of  an online judge system. It is shown that the model performed well in predicting the right category with high accuracy.
Intisar et al. \cite{Intisar} tried to classify the programming problems. For this, they made use of the two topic modeling techniques namely Non-negative Matrix Factorization (NMF) and Latent Dirichlet Allocation (LDA) for extracting the relevant features. Then, by utilizing these topic modeling features and Naive TF-IDF features, six classifiers were trained.  It is found a series of beneficial trade-offs between the applied models in terms of dimensionality and accuracy.
\vspace{-3ex}

\section{Proposed System}\vspace{-2.5ex}

The proposed system gets started with the collection of dataset from online programming practice platform Codeforces using its public Application Programmable Interface (API). Then some pre-processing tasks had been performed on the collected data to convert them into sequences. Then some state-of-the art sequence to  sequence models had been trained and tested on the collected data. 
\vspace{-5ex}
\subsection{Dataset Collection}
\vspace{-1ex}
Our proposed framework includes two phases. At first, we collected the data of 100 contestants from Codeforces using codeforces public API. The data\footnote{https://cutt.ly/nL120M9} includes contestant’s ratings, competition ranks, problem submissions, and submission verdicts. After collecting the data, we did some pre-processing. We considered each contest as a timestamp. Each timestamp has four types of features.
\begin{enumerate}
    \item Rating: Each contest represents a timestamp. The rating is a metric to evaluate an user/contestant. The more the rating is, the better performer the contestant is. This changes after each contest based on the rank of the user in that contest. We are going to predict this feature.
    \item Rank: This is the rank/position of the contestant in that contest. The rank is decided by the solve rating of the contestant in that contest. The more the solve rating is, the better rank the contestant gets.
    \item Solve Rating: Each contest has several problems and each problem has a different point based on its difficulty. The point of each problems decreases with time. The quicker a contestant solves a problem, the better points he gets. A contestant's solve rating is calculated by adding the points of each problem he solved during that competition. 
    \item Practice Features: This is the information of a contestant about how much practices he did after the previous contest and before the current contest.
        \begin{enumerate}
            \item 	Accepted(AC): It represents the number of problems that the contestant solved before the current contest and after the previous contest.
            \item Wrong Answer(WA): It denotes the number of problems that were attempted by the contestant but failed to solve it correctly before the current competitions and after the previous competitions.
        \end{enumerate}
\end{enumerate}
\vspace{-2ex}
Finally, we built a dataset of sequences where each continuous 16 timestamps of a user are considered as a sequence. Among the 16 timestamps, first 15 timestamps were used as the input and 16th timestamp was used as the target. 80\% of the sequences were used to train the models and the rest of the 20\% of the sequences were used to test/evaluate the performance of the trained models.

\vspace{-3ex}
\subsection{Frameworks}
\vspace{-2ex}
In the second phase of our proposed system, we apply several state-of-the-art neural network models to predict the performance of each contestant in the impending programming contests based on previous contests. First, we describe the concepts of Recurrent Neural Network (RNN), since Long Short Term Memory (LSTM), and Gated Recurrent Unit (GRU) both are categorized into that one. Then we describe Bidirectional LSTM (Bi-LSTM), and a combination of LSTM with an attention layer (LSTM+AL).

\vspace{-3ex}

\subsubsection{a. Recurrent Neural Network} RNN is a special class of ANN, which was originally proposed by Hopfield \cite{hopfield1982neural}. There is a basic difference between the conventional simple feed-forward neural network and RNN. Whereas in a feed-forward network, information flows in a single direction from the input nodes to the output nodes via the hidden nodes, RNN remembers the past sequences as well as being operated by the present node i.e. the system comes back to its previous node while running the current note. As a result, cycles or loops happen in the network. 
As it visits its previous nodes in every iteration, these RNN approaches perform well in sequence tasks and are widely used in prediction tasks e.g. stock market prediction, language translation, etc.

\vspace{-2ex}
\subsubsection{b. Long Short-Term Memory}
As mentioned earlier, RNN remembers the past sequences and puts on the proper context. Moreover, it remembers that information for a small duration of time. As a result, RNN falls short in terms of long sequences of data needed to process. Long Short Term Memory, which is commonly known as LSTM is a particular type of RNN, proposed by Hochreiter et. al. in 1997 \cite{hochreiter1997long}, which can mitigate this issue. While putting new information RNN transforms the existing information once applied a function. As a result, the entire information gets modified, on the whole, it fails to infer any such consideration for important or less important information. On the other hand, LSTM makes little modification to the information. In LSTM, this information flow is called cell states. In this way, LSTMs can selectively remember or forget things as per the context. The LSTM Architecture varies a little in terms of its internal components. Unlike RNN, it contains four internal cells inside a single LSTM block. In order to build the LSTM model, we used four LSTM layers with 256 neurons. After each LSTM layer, we used a dropout layer with a drop rate of 0.5. Then, we added a dense layer of 100 neurons with the activation function Relu. At last, a dense layer was used to output the features of the next timestamp of the sequence.

\vspace{-3ex}
\subsubsection{c. Bi-LSTM} 
\begin{figure}[!h]
    \centering
    \includegraphics[scale=.2]{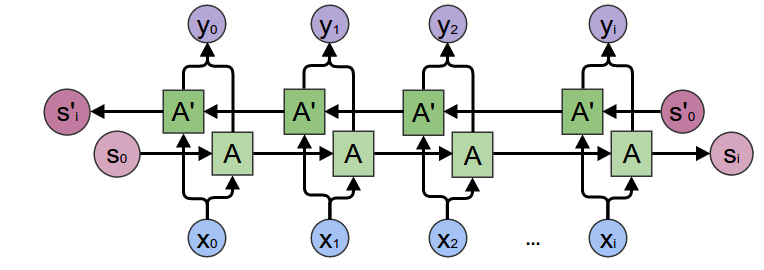}
    \vspace{-3ex}
    \caption{Bi-Directional LSTM block}
    \label{fig: bi-lstm}
    \vspace{-5ex}
\end{figure}

In RNN, Bidirectional LSTM is commonly known as Bi-LSTM where Bidirectional RNN are just putting two independent RNNs together. Similarly, Bidirectional LSTM is putting two independent LSTMs together so that the networks can have both backward and forward information about the sequence at every time stamp. Bi-LSTM processes inputs in both the past-to-future and future-to-past directions. The thing that differentiates this approach from the unidirectional one is LSTM runs backward it's preserved information from the future and uses the two hidden states combined which is able to preserve information from both the past and future at a given time. The simple building block of bidirectional LSTM has been shown in Fig. \ref{fig: bi-lstm}. \vspace{-5ex}

\vspace{-4ex}\subsubsection{d. LSTM with Attention Mechanism}
The Attention Mechanism is one of the most widely used methods in the Deep Learning research area. It was first proposed by Bahdanau et. al. in 2014 \cite{bahdanau2014neural}. The main bottleneck of the earlier methods of Attention Mechanism such as encoder-decoder-based RNNs/LSTMs, fall apart to deal with long sequences. Moreover, those fail to emphasize any important sequences or patterns. Then, the idea of Bahdanau et. al. was not only to keep track of long sequences but also to put more weight on the patterns which would be much needed to predict the outcome.

\vspace{-3ex}
\subsection{Models Configuration}
\vspace{-1.5ex}
In this paper, for all the models described above, we have configured four corresponding model layers with 256 neurons. After each layer, we used a dropout layer of drop rate 0.5. Then, we added a dense layer of 100 neurons with activation function Relu. Finally, a dense layer was used to output the features of the next timestamp of the sequence. In order to train the models, we used ‘mae’ and ‘adam’ as the loss function and optimizer respectfully. We trained all of the models for 1000 epochs with batch size 256. During training, weights of the best accuracy for each of the models were saved using the checkpoint. To check the efficacy of all the models with the test dataset, we used the saved weights of each of the trained models. The test dataset was sent through each of the models and accuracy was calculated to evaluate the models.
\vspace{-3ex}

\section{Experiment and Result Discussion}
\vspace{-2ex}
\subsection{Evaluation Metrics}
\vspace{-1.3ex}
Metrics such as  \textbf{Mean Absolute Error (MAE)} and  \textbf{Root Mean Squared Error (RMSE)}, which are often used to determine correctness for continuous data, have been employed to evaluate our proposed framework. These two metrics have been increasingly utilized by researchers to demonstrate the efficacy of their method \cite{Jaidka}, \cite{Bermingham}, \cite{Das}.
\par \textbf{Mean Absolute Error \textit{(MAE)}:} It is a terminology that provides the measurement of flaws in the assessment compared with original values. It is also referred to \textbf{Absolute Accuracy Error (AAE)}. \textbf{MAE} represents the average of all \textbf{AE}s. The MAE is denoted as, \vspace{-3ex}
\begin{equation}\label{eqn:MAE}
    MAE(y,y')=\frac{1}{N}\sum_{k=1}^{N}{\bigg| y_k - y'_k\bigg|}
\end{equation}
\vspace{0ex}
Here, $y_k$ and $y'_k$ denote the actual rating of $k^th$ sample and the predicted rating of them from the models mentioned respectively. $N$ refers the number of samples.

\par\textbf{ Root Mean Squared Error \textit{(RMSE)}:} It is a measurement of standard deviation that indicates how far the predicted value deviates from the actual value. Typically, this approach is suitable for finding the residuals' standard deviation. Residuals are the prediction errors, or the distance between the regression line and the actual data points.
Equation \ref{eqn:RMSE} shows how \textbf{RMSE(t)} is calculated. Where the notations are the same as described in Equation \ref{eqn:MAE}.
\vspace{-3ex}
\begin{equation}\label{eqn:RMSE}
    RMSE(y,y')=\sqrt[]{\frac{1}{N}{\sum_{k=1}^{N}{\bigg[ y_k - y'_k\bigg]^2}}}
\end{equation}
\vspace{-3ex}
\par\textbf{ Mean Squared Error \textit{(MSE)}:} It is also known as Mean Squared Deviation (MSD), which is another well known evaluation metric userd for prediction tasks.\vspace{-2.5ex}
\begin{equation}\label{eqn:MSE}
    MSE(y,y')={\frac{1}{N}{\sum_{k=1}^{N}{\bigg[ y_k - y'_k\bigg]^2}}}
\end{equation}
\vspace{-3ex}

\par \textbf{R-squared (\textbf{$R^2$}):} R-squared is a statistical metric that represents the proportion of the variance for an observed variable that's explained by a predicted variable or variables in a predictive model. The correlation describes the strength of the association between the observed and predicted values. It also describes the degree to which the variance of one variable explains the variation of the second variable. \textbf{R-squared} measure is ranged between $0$ to $1$ and usually mentioned as percentages. The more the value of this metric is, we consider the more precise the predictive model to be. Equation \ref{eqn:r_sq} shows how \textbf{R-squared} is calculated \cite{Biswas}.
\vspace{-2ex}
\begin{equation} \label{eqn:r_sq}
    R^2(y,y')=1- \frac{\sum_{k=1}^{N} {[y_k - y'_k]^2} }{\sum_{k=1}^{N}{[y_k - y' ]^2}}
\end{equation}
\vspace{-3ex}

\begin{equation} \label{eqn:r_sq1}
    y'= \frac{1}{N} \sum_{k=1}^N y_k
\end{equation}

\vspace{-5ex}
\subsection{Experimental Results}
\vspace{-2ex}
Does considering contestants' practice as an input feature helpful for better accuracy during the contestants' performance prediction? To answer this question, at first, we ran the models without the contestant's practice feature. Then, we ran the models without the contestant's practice features.
Result of the different models to predict the performance of the programmer on Codeforces is presented in Table \ref{tab:exp_results1} and Table \ref{tab:exp_results2}.

\begin{table}[ht]
\centering
\vspace{-2ex}
\caption{Experiments results of different models for different evaluation metrics without considering contestant’s practice features.}
\vspace{-2ex}
\begin{tabular}{|c|c|c|c|c|}
\hline
\textbf{\begin{tabular}[c]{@{}c@{}}Performance\\ Metrics\end{tabular}} & \textbf{LSTM}  & \textbf{\begin{tabular}[c]{@{}c@{}}LSTM + Attention\end{tabular}} & \textbf{GRU} & \textbf{Bi-LSTM} \\ \hline \hline
\textit{{RMSE}}                                                 & 73.133         & \textbf{69.629}                                                      & 85.968       & 75.024           \\ \hline
\textit{{MSE}}                                                  & 5348.436       & \textbf{4848.302}                                                    & 7390.642     & 5628.728         \\ \hline
\textit{{MAE}}                                                  & 57.523          & \textbf{54.243}                                                      & 66.324       & 59.069           \\ \hline
\textit{\textbf{$R^2$}}                                 & {0.906} & \textbf{0.930 }                                                               & 0.40        & 0.7942            \\ \hline
\end{tabular}

\label{tab:exp_results1}
\end{table}

\vspace{-7ex}

\begin{table}[ht]
\centering
\caption{Experiments results of different models for different evaluation metrics considering the contestant’s practice features.}
\vspace{-2ex}
\begin{tabular}{|c|c|c|c|c|}
\hline
\textbf{\begin{tabular}[c]{@{}c@{}}Performance \\ Metric\end{tabular}} & \textbf{LSTM}  & \textbf{\begin{tabular}[c]{@{}c@{}}LSTM + Attention\end{tabular}} & \textbf{GRU} & \textbf{Bi-LSTM} \\ \hline \hline
\textit{{RMSE}}                                                 & 59.287        & \textbf{51.325}                                                      & 72.342       & 62.244           \\ \hline
\textit{{MSE}}                                                  & 3948.436       & \textbf{3243.234}                                                    & 6089.834     & 4467.907        \\ \hline
\textit{{MAE}}                                                  & 42.67         & \textbf{39.217}                                                      & 53.219      & 45.989          \\ \hline
\textit{\textbf{$R^2$}}                                 & {0.946} & \textbf{0.97 }                                                               & 0.884       & 0.928            \\ \hline
\end{tabular}

\label{tab:exp_results2}

\end{table} 

\vspace{2ex} Table 2 presents the performance of different models for different evaluation metrics without considering the contestant’s practice features. Among all the four models, the LSTM with attention achieved the lowest RMSE value (69.629) outperforming other three models: LSTM (73.133), GRU (85.968), and Bi-LSTM (75.024). The LSTM with attention model also outperformed other three models in terms of other three evaluation metrics: MSE, MAE and $R^2$. On the other hand, when the contestant’s practice information is used as a feature, the performance of all models significantly improves (Table 3). The values of RMSE, MSE, and MAE decrease, and $R^2$ increase significantly for all models. From the above two tables, we can see that the LSTM with Attention model provides better accuracy than others in both cases. By analyzing the four-evaluation metrics, it is observed that LSTM with the Attention model performed better than the other three applied models, and GRU performed worst among the applied four models.  So, the sequence (from best to worst) of the performance of the models is LSTM with the Attention model, LSTM model, Bi-LSTM model, and GRU model. \vspace{-4ex}

\section{Discussion}
\vspace{-3.5ex}
In this work, we showed how contestants’ future performance could be predicted by employing deep learning models. 
We also found that using previous details practice information as input features improves the model’s accuracy significantly. When the practice information is included as an input feature, LSTM with Attention performs best. The RMSE, MSE, MAE and $R^2$ of LSTM with the Attention model were 51.325, 3243.234, 39.217, and 0.97 respectively (Table 3). Then the second-best model is LSTM which got RMSE 59.287, MSE 3948.436, MAE 42.67, and $R^2$ 0.946. We can see that LSTM with Attention provides at least $8\%$ better performance in every metric than the second-best model.
On the other hand, LSTM with Attention also achieved the highest efficacy excluding the practice information as the input feature as per our experiment. The RMSE, MSE, MAE, and $R^2$ of LSTM with the Attention model were 69.629, 4848.302, 54.243, and 0.93 respectively while the values of the second-best model LSTM are 73.133, 5348.436, 57.523 and 0.906 respectively (Table 2). LSTM with Attention shows at least $5\%$ better performance in every metric than the second-best model.
We found LSTM with Attention as the best model in both cases. We can see that the performance of LSTM with the Attention model improves significantly when the practice information is used as the input features. The RMSE, MSE, MAE, and $R^2$ values of LSTM with Attention model improve about $26\%$, $33\%$, $24\%$, and $10\%$ respectively.  It proves that the more a contestant practices, the better he/she does well in the future. In other words, we can conclude that practices make a big difference in the upcoming competitions. Therefore, without practice information, the prediction is not as accurate as when the prediction is done with practice information. Here, we collected the practice information only from CodeForce’s website. What if the contestant practices on other platforms or offline? In that case, our models will fall short to provide this performance. In the real world, these participants  can get alert about their signs of progress by the predictions of our proposed method and they can improve their skills to perform well in their next contests. In the future, we are planning to include practice information from other platforms too. Besides, we calculated the practice information by summing the number of problems solved before the contest whereas we didn’t consider the difficulty level of the problems. What if the contestant solves only the easier problems? It won’t help him to perform better but the model will predict that he will do better. 

\vspace{-3ex}
\section{Conclusion and Future Work}
\vspace{-3ex}
In this research, we provide a method for predicting participant ratings and analyzing their performance. We used a real-world Codeforces dataset to validate our methodology. The experiments had been conducted by considering the contestants' practice features as well as without considering them. In the future, we aim to consider each problem’s (solved before the next competition) difficulty level as a feature and employ the data from other platforms also.
\vspace{-4ex}

%
%
%
%

\end{document}